\documentclass[10pt, journal]{IEEEtran}
\usepackage{algorithm}
\usepackage{algorithmic}
\usepackage{amssymb} 
\usepackage{amsmath} 
\usepackage{multirow}
\usepackage{adjustbox}  
\usepackage{tabularx} 
\usepackage{newfloat}
\usepackage{listings}
\usepackage{xcolor}
\usepackage{url}
\usepackage[colorlinks=false, pdfborder={0 0 0}]{hyperref}

\let\oldurl\url
\renewcommand{\url}[1]{\textcolor{green!50!black}{\oldurl{#1}}}

 \makeatletter
    \newcommand{\thickhline}{%
        \noalign {\ifnum 0=`}\fi \hrule height 1pt
        \futurelet \reserved@a \@xhline
    }
    \newcolumntype{"}{@{\vrule width 1pt}}
    \makeatother

\ifCLASSINFOpdf
\else
\fi

\begin{document}

\title{PAMD: Plausibility-Aware Motion Diffusion Model for Long Dance Generation}

\author{Hongsong~Wang, Yin~Zhu, Qiuxia Lai, Yang Zhang, Guo-Sen Xie, and Xin~Geng,~\IEEEmembership{Senior~Member,~IEEE}
\thanks{H. Wang, Y. Zhu and X. Geng are with School of Computer Science and Engineering, Key Laboratory of New Generation Artificial Intelligence Technology and Its Interdisciplinary Applications, Ministry of Education, Southeast University, Nanjing 210096, China (\{hongsongwang, zhuy, xgeng\}@seu.edu.cn).}
\thanks{Q. Lai is with State Key Laboratory of Media Convergence and Communication, Communication University of China (qxlai@cuc.edu.cn).}
\thanks{Y. Zhang is with School of Computer Science and Software Engineering, National Engineering Laboratory for Big Data System 	Computing Technology, Guangdong Key Laboratory of Intelligent Information Processing, Shenzhen University, Shenzhen 518060, China (yangzhang@szu.edu.cn).}
\thanks{G. Xie is with School of Computer Science and Engineering, Nanjing University of Science and Technology, Nanjing, China (gsxiehm@gmail.com).}
}

\maketitle

\begin{abstract}
Computational dance generation is crucial in many areas, such as art, human-computer interaction, virtual reality, and digital entertainment, particularly for generating coherent and expressive long dance sequences. Diffusion-based music-to-dance generation has made significant progress, yet existing methods still struggle to produce physically plausible motions. To address this, we propose Plausibility-Aware Motion Diffusion (PAMD), a framework for generating dances that are both musically aligned and physically realistic. The core of PAMD lies in the Plausible Motion Constraint (PMC), which leverages Neural Distance Fields (NDFs) to model the actual pose manifold and guide generated motions toward a physically valid pose manifold. To provide more effective guidance during generation, we incorporate Prior Motion Guidance (PMG), which uses standing poses as auxiliary conditions alongside music features. To further enhance realism for complex movements, we introduce the Motion Refinement with Foot-ground Contact (MRFC) module, which addresses foot-skating artifacts by bridging the gap between the optimization objective in linear joint position space and the data representation in nonlinear rotation space. Extensive experiments show that PAMD significantly improves musical alignment and enhances the physical plausibility of generated motions. This project page is available at: \url{https://mucunzhuzhu.github.io/PAMD-page/}.
\end{abstract}


%
\IEEEpeerreviewmaketitle
\section{Introduction}
\IEEEPARstart{D}{ance} is an art form that harmonizes rhythmic body movements with musical accompaniment. It allows for the expression of emotions~\cite{sawada2003expression}, the preservation of cultural heritage~\cite{georgios2018transformation}, and the promotion of social connections~\cite{fink2021evolution}. However, traditional choreography is complex and labor-intensive, often requiring extensive professional expertise. Recent advances in computational methods provide an innovative alternative, pushing beyond the limitations of manual approaches by automatically generating dance movements synchronized with music. These methods can not only inspire new forms of artistic expression, but also offer immersive interaction experiences when paired with VR or AR technologies~\cite{tseng2023edge}. Furthermore, computational music-to-dance generation fosters interdisciplinary collaboration among computer science, performing arts, and human-computer interaction~\cite{shiratori2006dancing,kitsikidis2015game}.

A key challenge of computational music-to-dance generation lies in generating realistic and physically plausible dance sequences over long durations. Early autoregressive music-to-dance methods~\cite{li2021ai,kim2022brand,siyao2022bailando} generate dance movements iteratively by predicting each frame based on previously generated frames and corresponding music features. These methods, however, suffer from error accumulation over time, leading to issues like motion freezing~\cite{tseng2023edge} and an inability to generate plausible long sequences. In contrast, diffusion-based models~\cite{tseng2023edge} generate entire dance sequences at once, partially mitigating the problem of error accumulation. Despite this advantage, both autoregressive and diffusion-based methods share a significant limitation: they rely on a latent space to represent motion, yet this latent space is not specifically designed to guarantee physically plausible poses. As a result, both approaches struggle to generate realistic and physically plausible outputs.

\begin{figure}[tb]
	\centering
	\includegraphics[width=0.95\linewidth]{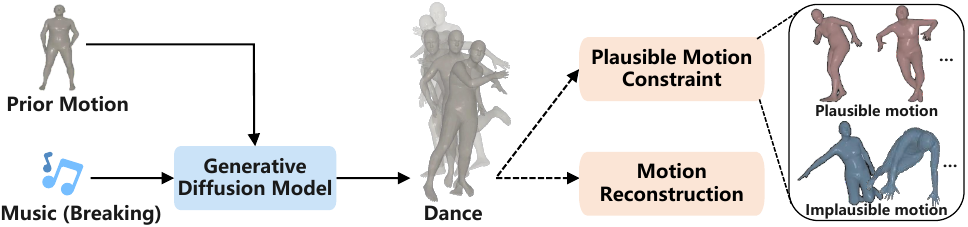}
	\caption{\textbf{Motivation of our approach for diffusion-based music-to-dance generation.} To generate plausible and correct motion sequences for music-to-dance, we introduce prior motion and plausible motion constraints during the training of the generative diffusion model.}
	\label{fig:intro1}
\end{figure}

\begin{figure*}[thb]
	\centering
    \includegraphics[width=0.95\linewidth]{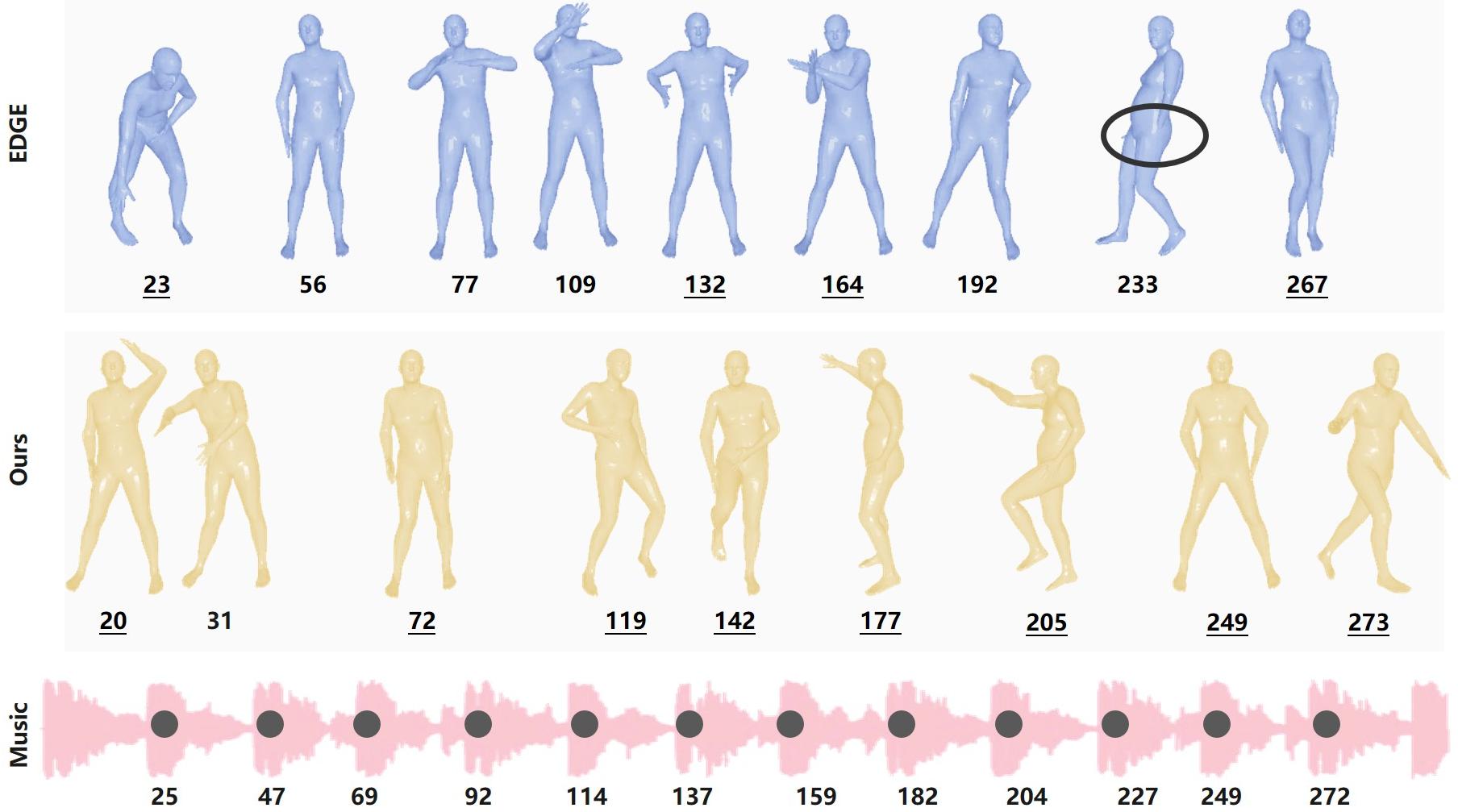}
   \caption{\textbf{Our PAMD model generates long dances that are better synchronized and visually coherent.} The black dots on the pink music waveform indicate music beats, while the grey and blue motions denote dance beats generated by PAMD (ours) and EDGE, respectively. The underlined dance beats indicate close alignment, which falls within five frames of the nearest music beat. PAMD produces eight closely aligned dance beats compared to only four from EDGE. Moreover, PAMD generates more natural and fluid movements. For example, EDGE lacks hand motion in frame 233, while PAMD maintains consistent and expressive hand movements in frame 205.}
    \label{fig:intro}
\end{figure*}

In general, seed motion is commonly utilized in autoregressive models, where the model predicts subsequent motions based on observed preceding motions~\cite{li2021ai}.
Although the seed motion renders the predicted motions physically plausible in short-term prediction, prediction errors of these models may accumulate for long-term dance prediction. Besides, many existing methods, such as \cite{li2021ai} and \cite{huang2022genre}, rely on auto-regressive inference to produce long dance sequences. This paradigm of long-term generation is inefficient and also leads to error accumulation. 

To address the limitations outlined earlier, this paper proposes Plausibility-Aware Motion Diffusion (PAMD), a diffusion-based framework for generating dances that are both musically aligned and physically realistic. Figure~\ref{fig:intro1} illustrates the outline of the proposed method.
PAMD consists of three key components, namely, Plausible Motion Constraint (PMC), Prior Motion Guidance (PMG), and Motion Refinement with Foot-Ground Contact (MRFC).
\textit{PMC}, inspired by human pose prior modeling~\cite{tiwari2022pose}, employs Neural Distance Fields (NDFs) to represent plausible human poses in a continuous, high-dimensional space. This constraint ensures the generation of physically valid motions by guiding outputs toward a realistic pose manifold. To the best of our knowledge, this is the first work to leverage NDFs for signal-conditioned human motion generation tasks, such as music-to-dance synthesis.
\textit{PMG} further enhances dance generation by providing a stable starting point for sequence generation, utilizing standing poses as auxiliary conditions. 
While conceptually similar to seed motions~\cite{rempe2021humor,duan2022unified} for predicting future movements, PMG does not depend on real motion data. Instead, it seamlessly integrates standing poses commonly seen in transitions, beginnings, or endings of dance sequences with music features, a strategy that has proven highly effective in our experiments.
\textit{MRFC} is inspired by~\cite{li2024lodge} and designed to improve the realism of complex movements by bridging the gap between the optimization objective in linear joint position space and the data representation in nonlinear rotation space. Unlike~\cite{li2024lodge}, MRFC is more lightweight, avoiding multiple training stages and reducing the computational costs, while still achieving comparable refinement quality.
As shown in Figure~\ref{fig:intro}, these components enable PAMD to generate high-quality, plausible dance sequences efficiently and effectively.
Our contributions are summarized as follows:
\begin{itemize}
	\item We introduce Plausibility-Aware Motion Diffusion (PAMD), a diffusion-based framework for music-to-dance generation. PAMD generates dances that are both musically aligned and physically realistic.  
	
	\item To ensure realistic dance generation, we design Plausible Motion Constraint (PMC), which uses NDFs to model plausible human poses on a continuous manifold, the first application of NDFs in music-to-dance generation.  
	
	\item To provide a stable starting point for sequence generation, we introduce Prior Motion Guidance (PMG), which uses standing poses as auxiliary conditions. 
\end{itemize}

\section{Related Work}

\subsection{Music-Driven Dance Generation}
Music-driven dance generation has become increasingly popular recently. Early approaches~\cite{ofli2011learn2dance,fan2011example,berman2015kinetic,lee2013music} generate dance by viewing this task as a motion retrieval problem. However, these approaches are limited by the quality and diversity of motion databases, lack flexibility in adapting to new music, and often produce mechanically repetitive movements.

Subsequently, deep learning approaches have gradually been applied to music-driven dance generation task, including autoregression-based methods~\cite{li2021ai, valle2021transflower, huang2022genre}, GAN-based methods~\cite{kim2022brand,sun2020deepdance}, VAE-based methods~\cite{siyao2022bailando,gong2023tm2d}, and Diffusion-based methods~\cite{tseng2023edge,li2024lodge,zhang2024bidirectional,luo2024popdg,huang2024beat}. 
Li et  al.~\cite{li2021ai} learn the correspondence between music and motion with a cross-modal transformer block. 
Apart from the music itself, Huang et  al.~\cite{huang2022genre} embed the genre one-hot vector into the decoder based on a transformer-based architecture.
However, these methods generate dance sequences in an autoregressive manner, which may lead to error accumulation.
Based on GAN, Kim et  al.~\cite{kim2022brand} use a transformer-based conditional GAN with a genre discriminator to generate diverse dance motions. Based on VQ-VAE, Li et al.~\cite{siyao2022bailando} achieve alignment between music beats and motion tempos with an actor-critic-based reinforcement learning scheme. Liang et al.~\cite{liang2024dancecomposer} propose a framework that establishes rhythmic and stylistic correlations between dance and music.
Recently, there have been a few diffusion-based methods for dance generation. 
Li et  al.~\cite{li2024lodge} propose a two-stage coarse-to-fine diffusion architecture, that first generates characteristic dance primitives and subsequently utilizes these primitives to generate longer dance sequences.
Zhang et  al.~\cite{zhang2024bidirectional} focus on local and bidirectional motion and employs a bidirectional encoder to process forward noise distributions and backward dance sequences.

Although deep learning-based methods have advanced the development of the music-to-dance generation task, they share a notable limitation: they rely on a latent space to encode motion, yet this latent space is not explicitly optimized to guarantee physically plausible poses. In contrast, our method leverages Neural Distance Fields (NDFs) to model the motion directly in the physical space. This approach ensures that the generated poses are not only more realistic but also physically plausible, as the NDF framework constrains the motion to adhere to natural physical laws. Therefore, our method overcomes the limitations of traditional latent space representations and enhances the overall plausibility of the generated poses.

\subsection{Conditional Human Motion Generation}
Most research within the human motion generation field focuses on generating human motions based on conditional signals~\cite{zhu2023human}. The conditional signals are usually class~\cite{guo2020action2motion,petrovich2021action}, text prompts~\cite{zhang2024motiondiffuse,petrovich2022temos}, audio features~\cite{ao2023gesturediffuclip}, observed motions~\cite{shi2023multi}, or scene contexts~\cite{ling2020character}.
Conditioned on class information, Guo et  al.~\cite{guo2020action2motion} propose a novel VAE framework to iteratively generate human motion sequences given a prescribed action type. 
Based on text prompts, Zhang et  al.~\cite{zhang2024motiondiffuse} propose the first diffusion-based framework conditioned on text descriptions with multi-level manipulation.
Considering audio styles,  Ao et  al.~\cite{ao2023gesturediffuclip} propose a neural network framework that synthesizes stylized co-speech gestures with flexible style control.  Ling et  al.~\cite{ling2020character} combine reinforcement learning with motion generative model to produce precise goal-directed movements under joystick control.

Conditioned on prior motions, many methods utilize prior motion information to guide and refine the generation or prediction of human motion.
Holden et  al.~\cite{holden2016deep} train a feedforward control network to generate realistic motion sequences based on terrain trajectory or a target location. Rempe et  al.~\cite{rempe2021humor} introduce a 3D human motion model and leverage this model as a motion prior for the prediction of temporal pose. Duan et  al.~\cite{duan2022unified} propose a trainable mixture embedding module to model temporal information, which controls the completion of motion. These approaches require a large amount of prior motion data for training, whereas our method only uses standing poses as auxiliary conditions commonly observed in transitions, beginnings, or endings of dance sequences, reducing computational and time costs significantly.

\subsection{Plausible Human Motion Generation}
Despite notable progress in recent years, the task of human motion generation remains challenging due to the complex nature of human movement and its implicit connection with conditional signals. Therefore, to prioritize the plausibility of the generated motions, Shimada et  al.~\cite{shimada2020physcap} utilize a combination of ground reaction force and residual force for plausible root control. Then, Shimada et  al.~\cite{shimada2021neural} propose a method called “physionical”, which is aware of physical and environmental constraints.  Yuan et  al.~\cite{yuan2021simpoe} combine a dynamics-based control generation unit with a kinematic pose refinement unit to achieve plausible pose generation. In addition to considering kinematics, Zhang et  al.~\cite{zhang2024physpt} develop a physics-driven body representation and a contact force model. Huang et  al.~\cite{huang2024closely} introduce a proxemics and physics-guided diffusion model, enabling the interaction to be modeled through cross-attention. Lodge~\cite{li2024lodge} uses a foot refine block, which extracts footstep information as an additional input condition to mitigate artifacts. However, this block exhibits a structure similar to that of a transformer and incurs considerable computational cost. To achieve a more refined understanding of foot-ground contact while minimizing computational overhead, we introduce a Motion Refinement with Foot-Ground Contact Module, which is not only lightweight but also effectively ensures the generation of plausible motions.

\section{Plausibility-Aware Motion Diffusion}
\begin{figure*}[thb]
	\centering
	\includegraphics[width=1\linewidth]{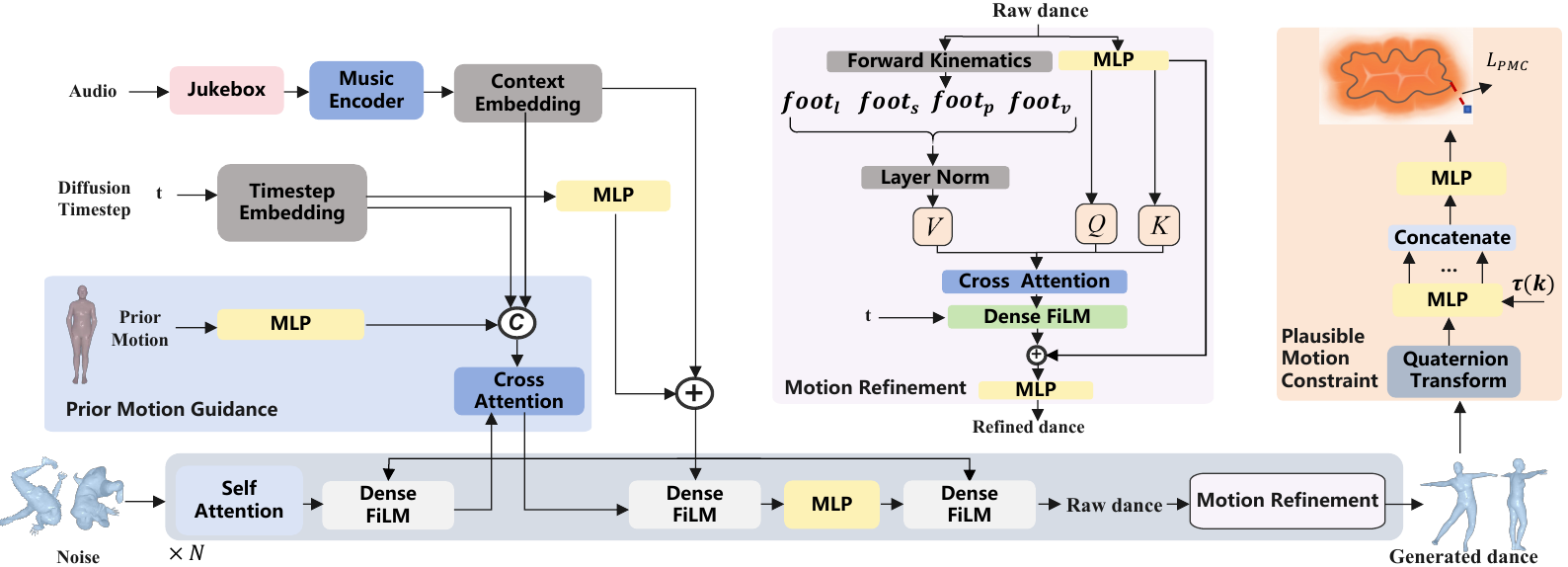}
	\caption{\textbf{PAMD Pipeline Overview:}  Conditioned on music and prior motion, PAMD learns to denoise dance sequences from time $t=T$ to $t=0$. Music features are extracted by Jukebox and then pass through the Transformer Music Encoder. The prior motion, timestep, and music features are concatenated and undergo cross-attention with noise. The noisy sequence $\hat{x}_t$ is processed by a transformer-based dance decoder, which generates the $raw$ $dance$. Motion Refinement Module takes $raw$ $dance$ as input, extracts $foot_{l}$, $foot_{s}$, $foot_{p}$ and $foot_{v}$, goes through a cross-attention, and outputs the final refined dance sequences. During the training process, the generated dance is passed through the Plausible Motion Constraint to produce an auxiliary loss. }
	\label{fig:model}
\end{figure*}

To pull the generated dance motions toward a physically plausible space, we propose a method called Plausibility-Aware Motion Diffusion (PAMD). This model instills physical principles of dances in multiple aspects, including Plausible Motion Constraint, Prior Motion Guidance and Motion Refinement with Foot-Ground Contact, and possesses the capability to generate long dances of arbitrary lengths. The pipeline overview of our method is shown in Figure~\ref{fig:model}. Before elaborating on our approach, we first present a baseline for music-to-dance generation as preliminaries.

\subsection{Preliminaries}
We refer to EDGE~\cite{tseng2023edge} and use Motion Diffusion Model (MDM)~\cite{tevet2022human} for dance generation. For music features, we use Jukebox to extract music features as conditions. Given a music clip, music features can be represented as $m \in {\mathbb{R}}^{L \times 4800} $, in which $L$ is the frame number and 4800 is the music feature channel. For motion representation, we obey the SMPL~\cite{loper2015smpl} format and employ the 6 degrees of freedom rotation representation~\cite{zhou2019continuity}  for each joint along with a single root translation: $\omega\in \mathbb{R}^{24\cdot6 + 3 =147} $. We additionally incorporate a binary contact label for both the heel and toe of each foot:  $ f \in {\{0,1\}}^{2\cdot2 = 4} $. Therefore, the total motion representation for each time step is $ x = \{f,w\} \in \mathbb{R}^{4+147=151} $.

The motion diffusion model is composed of two main processes: a forward noise addition process and a reverse denoising process. The forward noise addition process is described as a Markov noising process:
\begin{align}
	q(x_t|x_0) = \sqrt{\overline{\alpha}_{t}}x_0 + (1-\overline{\alpha}_{t})\epsilon,
\end{align}
where $x_0$ is ground truth dance data, $\epsilon \sim \mathcal N (0,\boldsymbol{I}) $ and $ \overline{\alpha}_t \in (0,1)$ are constants that are monotonically decreasing. The forward process aims to perturb $x_0$ into $x_t$ over $t$ steps. 

The reverse process mainly uses a network $g_\theta$ to recover the motion from noise, generating $\hat{x}_0$ conditioned on music $m$. We optimize $g_\theta$ using the reconstruction loss~\cite{ho2020denoising}:
\begin{align}
	{\mathcal L_{\text{recon}} = {\mathbb{E}}_{x_0,t}[{||x_0-g_\theta(x_t,t,m)||}^2_2]}.
	\label{eq:recon}
\end{align}
where $\hat{x}_0 = g_\theta(x_t,t,m)$ and $m$ denotes the music features.

Following the approach~\cite{ho2022classifier}, we incorporate classifier-free guidance by introducing a low probability (e.g. 20\%) of randomly replacing $  c = \emptyset$ during training. 
The guided inference is formulated as:
\begin{equation}
	\widetilde{x}_0 = w \cdot g_\theta(x_t,t,m) + (1-w) \cdot g_\theta(x_t,t,\emptyset), 
\end{equation}
where $w$ is the guidance weight with a positive value. The influence of condition $ m $  can be amplified by setting $w>1$.

For human motion synthesis, auxiliary losses are commonly employed to enhance the physical realism of the generated motions~\cite{petrovich2021action}. We follow Tevet et al.~\cite{tevet2022human} to incorporate three auxiliary losses: joint position loss ${\mathcal L}_{\text{joint}}$, velocity loss ${\mathcal L}_{\text{vel}}$ and foot contact loss ${\mathcal L}_{\text{foot}}$:
\begin{align*} 
	&{\mathcal L}_{\text{joint}} = \frac{1}{N} \sum\limits_{i=1}^{N} {||\mathrm {FK}(x^{(i)}) - \mathrm {FK}({\hat{x}}^{(i)})||}^2_2,  \\
	&{\mathcal L}_{\text{vel}} = \frac{1}{N-1} \sum\limits_{i=1}^{N-1}{||(x^{(i+1)}-x^{(i)})-({\hat{x}}^{(i+1)}-\hat{x}^{(i)})||}^2_2, \\
	&{\mathcal L}_{\text{foot}} = \frac{1}{N-1} \sum\limits_{i=1}^{N-1}{||(\mathrm{FK}(\hat{x}^{(i+1)})-\mathrm{FK}({\hat{x}}^{(i)})) \cdot \hat{f}^{(i)}||}^2_2. 
\end{align*} 
where $\mathrm {FK}(\cdot)$ denotes the forward kinematic that converts joint angles to positions, $\hat{f}^{(i)}$ is the binary contact label for foot and the superscript $(i)$ indicates the frame index.


\subsection{Plausible Motion Constraint}
\begin{figure}[tb]
	\centering
	\includegraphics[width=1\linewidth]{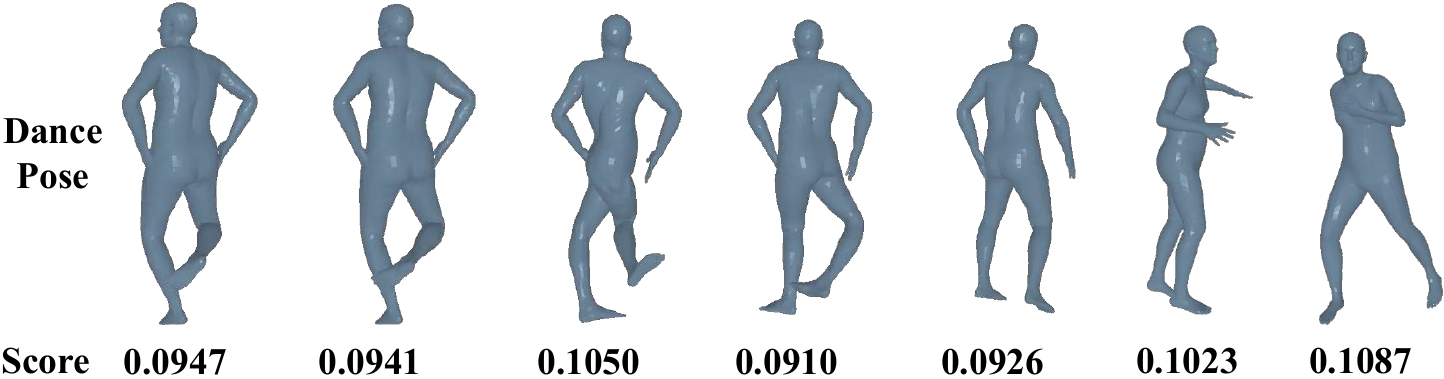}
	\caption{\textbf{Implausible poses of dance generation:} 
		The score indicates the implausibility of dance poses, with higher scores indicating less plausible poses. It is predicted by the trained auxiliary network in the PMC module.
	}
	\label{fig:NDF3}
\end{figure}

In diffusion models, motion is represented in a latent space, which makes it challenging to ensure that the generated motions are physically plausible. In the dance generation task, some typical implausible poses for the task of dance generation are shown in Figure~\ref{fig:NDF3}. These abnormal poses diminish the quality and visual appeal of the generated dance. 

Therefore, we introduce the Plausible Motion Constraint (PMC), which employs Neural Distance Fields (NDFs) to represent plausible human motions in a continuous, high-dimensional space to serve as a constraint in dance generation. This module aims to learn a neural network $f$ that maps a human pose $\theta \in SO(3)^K$ to a non-negative scalar, i.e.,  $f: SO(3)^K \rightarrow {\mathbb{R}}^+$. The manifold of plausible poses is represented as the zero-level set:
\begin{align}
	S = \{\theta \in SO(3)^K| f(\theta)=0\},
\end{align}
where  $SO(3)^K$ denotes the pose space, $K$ is the number of body joints, and the value of $f(\theta)$ signifies the unsigned distance from the pose $\theta$ to the manifold of plausibility.

The PMC module consists of an encoder $f^{\text{enc}}$ and a decoder $f^{\text{dec}}$, following the practice of 
modeling pose manifolds with neural distance fields~\cite{tiwari2022pose}. Given a pose $\theta = \{\theta_1, ..., \theta_K\}$, where $\theta_k$ is the pose for joint $k$, $f^{\text{enc}}$ encodes each human pose using an MLP as:
\begin{align}
	v_1 = f_1^{\text{enc}}{(\theta_1)}\notag,\\
	v_k = f_k^{\text{enc}}{(\theta_k, v_{\tau(k)})},
\end{align}
where $\tau(k)$ is a function that maps the index of each joint to the index of its parent joint. To model a continuous manifold of human poses, the quaternion transform~\cite{wang2021pvred} is employed first to map joint angles of rotation representations to unit quaternion representations. 

Given an input pose and the encoded feature of the parent joint for each joint in the pose, $f^{\text{enc}}$ outputs the encoded feature $v_i$ for each joint in the pose. Then these features are concatenated to form $V=[v_1||...||v_K]$. Subsequently, $V$ is fed into the decoder $f^{\text{dec}}$, which predicts the unsigned distance between the given pose and the corresponding plausible pose manifold using an MLP: 
\begin{align}
	d = f^{\text{dec}}(V),
\end{align}
where $d$ measures the motion plausibility, $d = 0$ indicates that the pose is plausible, and a larger value of $d$ signifies a more implausible pose.

The plausible motion constraint $\mathcal L_{\text{PMC}}$ is defined as:
\begin{align}
	\mathcal L_{\text{PMC}} = \frac{1}{N} \sum\limits_{i=1}^{N} f^{\text{dec}}\Big(f^{\text{enc}}\big( q(\hat{x}^{(i)})\big)\Big),
	\label{eq:loss_ndf}
\end{align}
where ${\hat{x}^i}$ is the generated motion of the $i$-th time step and $q(\cdot)$ denotes the function of quaternion transform. 
Compared to previous methods that transform human pose into Gaussian distributions~\cite{rempe2021humor}, the PMC module models the actual pose manifold, preserving distances between real poses.
Therefore, the total training loss is:
\begin{align}
	{\mathcal L} = {\mathcal L}_{\text{recon}}  & +  \lambda_{\text{joint}}{\mathcal L}_{\text{joint}} + \lambda_{\text{vel}}{\mathcal L}_{\text{vel}} \notag\\ & + \lambda_{\text{foot}}{\mathcal L}_{\text{foot}} +\lambda_{\text{PMC}}{\mathcal L}_{\text{PMC}}.
\end{align}

\subsection{Prior Motion Guidance}
Seed motion refers to the small number of initial frames provided, which are used to represent the start of human motion. FACT~\cite{li2021ai} and GCDG~\cite{huang2022genre} generate subsequent motions automatically based on given seed motions. It has been found that providing a seed motion can effectively guide the generation of subsequent short-term motions. 

\begin{figure}
	\centering
	\includegraphics[width=1\linewidth]{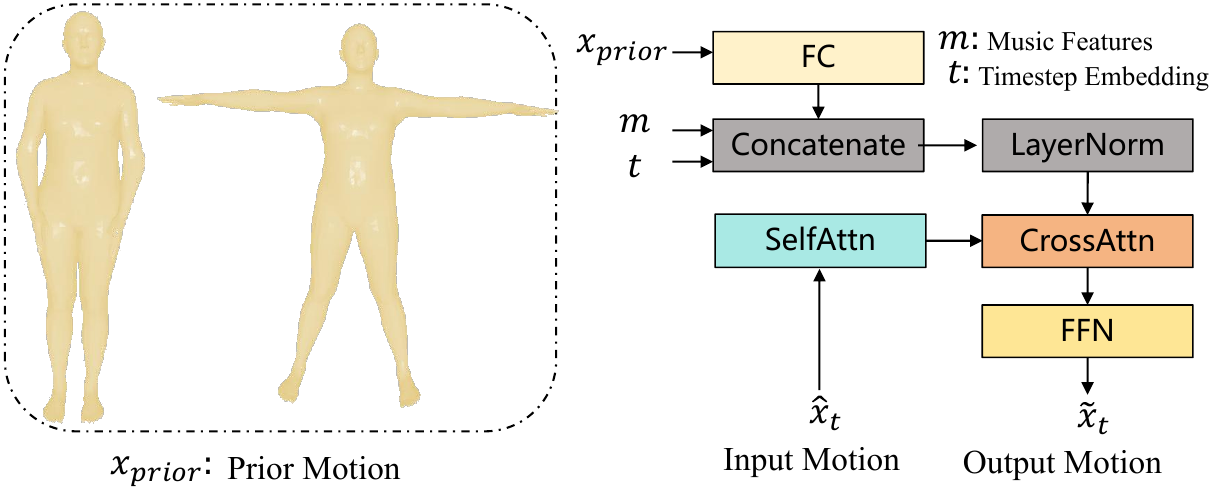}
	\caption{\textbf{Prior Motion Guidance:}   
		$x_{\text{prior}}$ is the chosen prior motion; $m$ and $t$ are music features and timestep token;  $\hat{x}_t$ is the input noisy dance sequence. $\tilde{x}_0$ denotes the output raw dance.
	}
	\label{fig:eq12}
\end{figure}

Based on the effectiveness of the seed motion, we design a prior motion as an additional condition to guide dance generation. 
Unlike the seed motion, which represents the beginning of a real motion, the prior motion signifies the prior knowledge of movement patterns, and we treat it as a fixed condition of the model rather than observed data. 
We choose a standard standing pose as the prior motion, as we observe that this pose or its slight variations are universally involved in nearly all dances, often serving as the beginning, end, or transition between movements. Specifically, the input noisy dance sequence, music features, timestep token and prior motion are denoted as $\hat{x}_t$, $m$, $t$ and $x_{\text{prior}}$, respectively. As shown in Figure~\ref{fig:eq12}, $\hat{x}_t$ first undergoes a self-attention module, then serves as $Q$ and $K$ in a cross-attention module. $m$, $t$ and $x_{\text{prior}}$ are projected into the same dimension and then concatenated as $V$ in the cross-attention module. Finally, it passes through a Feedforward Neural Network to output raw dance. 

\begin{algorithm}[tb]
	\caption{Long dance generation algorithm}
	\label{alg:long}
	\textbf{Input}: The conditioned music. \\
	\textbf{Parameter}: Diffusion time step $T$, the slice length $L$.\\
	\textbf{Output}: Generated long dance sequences $\tilde{x}$. 
	
	\begin{algorithmic}[1] 
		\STATE Slice the music, ensuring the latter half of previous slice is the same as the former half of next slice; get a total of $N$ slices; each slice has $L$ frames; $h = L/2$
		\STATE Extract features $m \in {\mathbb{R}}^{N \times L \times 4800}$ from music slices
		\STATE Generate a random noise $\hat{x}_T \in  {\mathbb{R}}^{N \times L \times 151} $
		\FOR{$t=T$ to $1$}
		\STATE Generate dance slices $\hat{x}_0 = \mathrm{PAMD}(\hat{x}_t)$
		\STATE Add noise to time step $t-1$ as $\hat{x}_{t-1} =q(\hat{x}_{t-1}|\hat{x}_0)$
		\STATE Assign the latter half of the previous dance slice to the former half of the next dance slice as: \\ $\hat{x}_{t-1}[1:, :h] = \hat{x}_{t-1}[:-1, h:]$
		\ENDFOR
		\STATE Set the weight $\beta = $ torch.linspace(1, 0, $h$) 
		\STATE Initialize dance $\tilde{x}$ of length $L + (N-1) \cdot h$ with zeros
		\FOR{$i=0$ to $N-1$}
		\IF{$i > 0$}
		\STATE $\hat{x}_0[i, 0:h] = \beta \odot \hat{x}_0[i, 0:h]$
		\ENDIF
		\IF{$i < N-1$}
		\STATE $\hat{x}_0[i, h:] = (1 - \beta) \odot \hat{x}_0[i, h:]$
		\ENDIF
		\STATE $\tilde{x}[h \cdot i : h  \cdot i + L] += \hat{x}_0[i]$
		\ENDFOR
	\end{algorithmic}
\end{algorithm}

\subsection{Motion Refinement with Foot-Ground Contact}
Compared to other body joints, foot joints can better reflect the quality of generated motion. Foot joints serve as leaf nodes in the SMPL model. If parent nodes rotate, such as the leg and knee, the foot joints may change a lot. 
However, most of the existing works of human motion generation only add a foot contact loss ${\mathcal L}_{\text{foot}}$. Lodge \cite{li2024lodge} employs a module aimed at eliminating artifacts. However, this module shares a similar structure with the dance decoder and is computationally expensive.

To gain deep insights into foot contact and avoid excessive computational overhead, we introduce a Motion Refinement with Foot-Ground Contact Module following the dance decoder block, as shown in Figure~\ref{fig:model}. Following ProxyCap~\cite{zhang2024proxycap}, this module first calculates joint positions through forward kinematics, then extracts the positions of the foot joints $f_{p}$, computes foot velocities $f_{v}$, and records foot contact labels $f_{l}$. Subsequently, it calculates the contact score $f_{s}$:
\begin{align}
	f_{s} = \mathrm{Sigmoid}(\frac{h_{\text{max}}-h_i}{k_h}) \cdot \mathrm{Sigmoid}(\frac{v_{\text{max}}-v_i}{k_v}),
\end{align}
where $h_i$ and $v_i$ denote the height and velocity of the given joint. Referring to the ProxyCap, we set $ k_h = 5 \cdot  h_{\text{max}} $ and $ k_v = 5 \cdot v_{\text{max}}$. 

Next, the raw dance generated by the dance decoder block is projected into latent dimension and serves as $Q$ and $K$ in a cross-attention module. $f_{l}$, $f_{s}$, $f_{p}$ and $f_{v}$ are concatenated and then projected into latent dimension to serve as $V$ in the cross-attention module. Finally, it passes through an FC to output the refined dance.

\subsection{Parallel Long Dance Generation}
Generating long dance sequences is essential in real applications. Long dance generation is challenging, requiring the model to maintain long-term dependencies in temporal patterns and transitions.
Many existing approaches employ auto-regressive inference to generate long motion sequences, which is not efficient for parallel computing. Instead, we introduce a method for parallel long dance generation.
Specifically, we first slice the music, ensuring the latter half of the previous music slice is the same as the former half of the next music slice. At each time step of the inference, PAMD first generates the dance slices according to music slices in parallel and then assigns the latter half of the previous dance slice to the former half of the next dance slice. Eventually, the dance slices are merged and the dance slices with the same music part are combined via weighted summation to generate the long dance. The whole inference process is presented in Algorithm~\ref{alg:long}.

\section{Experiments}

\begin{table*}[thbp]
	\centering
	\caption{Results of dance generation on the AIST++. $\uparrow$ means higher values indicate better performance, $\downarrow$ means lower values indicate better performance and $\rightarrow$ means closer to the ground truth is better. During inference, similar to EDGE~\cite{tseng2023edge}, the default setting for the guidance weight $w$ is 2. Additionally, results using $w=1$ are presented to investigate the effects under diminished conditions.}
	\vspace{-5pt}
	\adjustbox{max width=0.8\textwidth}{ 
		\begin{tabular}{llcccccc}
			\hline
			Method & Model & $BAS\uparrow $  & $PFC\downarrow$   & $FID_k\downarrow$ & $FID_g\downarrow$ & $Div_k\rightarrow$& $Div_g\rightarrow$\\
			\hline
			EDGE (w=2)~\cite{tseng2023edge}& Diffusion&  0.26    &    1.56  &   35.35    &   18.92    &    5.32   &  4.90 \\
			EDGE (w=1)~\cite{tseng2023edge}& Diffusion&   0.25   &    1.19 &   45.41   &   19.42    &   4.82   &  5.24 \\
			PAMD (w=2) (Ours) & Diffusion&  \textbf{0.31}    &   1.44    &  35.13     &  17.59     &    5.94  & 4.72 \\
			PAMD (w=1) (Ours) & Diffusion&  0.27   &  \textbf{1.03}      &  42.67     &   \textbf{17.46}  &   4.99   &  5.00 \\
			\hline
			FACT~\cite{li2021ai} & Others&      0.20   & 30.39   & 561.14  & 170.36   &--  & --  \\
			Bailando~\cite{siyao2022bailando} & VQ-VAE&       0.21  & 1.72  & 24.30& 20.81  & 6.83  & \textbf{7.69}  \\
			TM2D~\cite{gong2023tm2d}  &  VQ-VAE&    0.19  & 3.28  & \textbf{15.37 } & 28.35  & \textbf{9.10} & 7.91 \\
			BADM~\cite{zhang2024bidirectional} & Diffusion&  0.24 &  1.42 &  -- & -- & --  &  --  \\
			Lodge~\cite{li2024lodge} & Diffusion&    0.24  & --     & 37.09  & 18.79  & 5.58  & 4.85  \\
			\hline 
			Ground Truth & &0.35  & 1.33  & --     &  --    & 9.29  & 7.46 \\
			\hline
	\end{tabular}}
	\label{tab:results}
\end{table*}

\begin{table*}[htbp]
	\centering
	\caption{Results of long dance generation on the AIST++ dataset. 
		We choose EDGE~\cite{tseng2023edge} as the comparative method, and do not compare our method with other methods because they are not designed for long-term generation and their trained models are not released.}
	\vspace{-5pt}
	\adjustbox{max width=\textwidth}{ 
		\begin{tabular}{l|cccccc|cccccc}
			\hline
			\multirow{2}[0]{*}{Method} & \multicolumn{6}{c|}{Generation Length 7.5s}  & \multicolumn{6}{c}{Generation Length 10s} \\
			\cline{2-13}
			& $BeatAlign\uparrow $  & $PFC\downarrow$   & $FID_k\downarrow$ & $FID_g\downarrow$ & $Div_k\rightarrow$& $Div_g\rightarrow$ & $BeatAlign\uparrow $  & $PFC\downarrow$   & $FID_k\downarrow$ & $FID_g\downarrow$ & $Div_k\rightarrow$& $Div_g\rightarrow$ \\
			\hline
			EDGE (w=2)~\cite{tseng2023edge} & 0.26  & 1.11  & 59.43 & 25.44 & 2.93  & 3.52 & 0.25  & 0.89  & 68.31 & 32.04 & 2.52  & 3.13 \\
			EDGE (w=1)~\cite{tseng2023edge} &  0.25     & 1.05      &  58.84   &  23.13     & \textbf{3.16}      & 4.24 & 0.25     & 0.86      &  68.11   & 31.19    & 2.68      & 3.52 \\
			PAMD (w=2) &  \textbf{0.30}     & 1.16      &  \textbf{57.98}   &  21.37  & 3.12      & 3.77 & \textbf{0.31}    & 0.83      &  \textbf{67.23}   &   28.65    &     2.72  &  3.35\\
			PAMD (w=1)) &  0.26     & \textbf{1.04}     &  60.36  &  \textbf{20.21}     & \textbf{3.16}     & \textbf{4.35} &  0.27     & \textbf{0.81}      &  67.72  &  \textbf{28.36}     & \textbf{2.91}      & \textbf{3.81} \\
			\hline
			Ground Truth & 0.38  & 1.04  & --    & --    & 9.29  & 7.46  & 0.49  & 1.7   & --    & --    & 9.29  & 7.46 \\
			\hline
	\end{tabular}}
	\label{tab:long_results}
\end{table*}

		

\begin{table*}[thbp]
	\centering
	\caption{Results of ablation studies of our approach on the AIST++ dataset.
	}
	\vspace{-5pt}
	\adjustbox{max width=0.8\textwidth}{ 
		\begin{tabular}{c|ccc|cccccc}
			\hline
			Method & PMC & PMG   & MRFC &  $BAS\uparrow $  & $PFC\downarrow$   & $FID_k\downarrow$ & $FID_g\downarrow$ & $Div_k\rightarrow$& $Div_g\rightarrow$ \\
			\hline
			\normalsize{\textcircled{\scriptsize{0}}} &       &       &        &  0.27      &   2.04    &  43.72   &   21.05  & 4.53   & 4.24  \\
			\normalsize{\textcircled{\scriptsize{1}}} &  \checkmark   &       &       &    0.28    &    1.66  &    39.14  & 19.53    &    4.98& 4.46 \\
			\normalsize{\textcircled{\scriptsize{2}}} &  \checkmark   & \checkmark     &      &   0.29   &   1.62   &   36.35   &   18.79   &  5.12    & 4.60 \\
			\normalsize{\textcircled{\scriptsize{3}}} &  \checkmark      &       & \checkmark       &    0.29   &  1.56     &  35.49   &   18.25    &   5.03   & 4.68 \\
			\normalsize{\textcircled{\scriptsize{4}}} &  \checkmark       & \checkmark       & \checkmark     &   \textbf{0.31}    &    \textbf{1.44}  &  \textbf{35.13}     &    \textbf{17.59}  & \textbf{5.94}     &  \textbf{4.72} \\
			\hline
	\end{tabular}}
	\label{tab:ablations}
	\vspace{-5pt}
\end{table*}

\subsection{Experimental Setup}

\noindent\textbf{Dataset: }
We use the AIST++ dataset~\cite{li2021ai}, which contains 1408 dance sequences ranging from 7 seconds to 50 seconds. 
For short-term dance prediction, we follow the setting of EDGE~\cite{tseng2023edge} that uses 5-second clips at 30 FPS with a stride of 0.5 seconds for the training process. Instances in the testing set are also segmented into 5-second clips at 30 FPS with a stride of 2.5 seconds.

\noindent\textbf{Evaluation Metrics: }
Beat Alignment Score (BAS) measures the synchronization between the music and the generated dance, which is calculated as the average time distance between each music beat and its closest dance beat. The music beats are extracted using the Librosa~\cite{mcfee2015librosa} package, and the kinematic beats are derived from local minima in the velocity of motion joints.

Physical Foot Contact (PFC) score is a metric that can measure the realism of foot-ground contact~\cite{tseng2023edge}. Any generated dance must adhere to physical plausibility; otherwise, its practical application is severely limited.

Frechet Inception Distance (FID) represents the distance between the distribution of generated dance motions and that of real motions. It reflects the motion quality of generated dance sequences. We compute $FID_k$ in the kinetic feature space and $FID_g$ in the geometric feature space. $FID_k$ quantifies the physical reality of the motion, while $FID_g$ assesses the overall quality of the dance choreography.

Diversity (Div) evaluates the diversity of generated dance sequences by computing the mean Euclidean distances within the motion feature space. Likewise, we calculate the diversity metrics $Div_k$ and $Div_g$ for the kinetic feature space and the geometric feature space, respectively.

\noindent\textbf{Implementation Details: }
The proposed model has 207.44 million (M) parameters. 
The training batch size is 128, the number of epochs is 2000, the learning rate is 0.0004 and the weight decay is 0.02. To reduce the influence of stochastic randomness, we perform the test process 100 times and average the results of the evaluation metrics. 

\noindent\textbf{Settings of Long Dance Generation: }With few research focused on long dance generation, there is a lack of well-recognized benchmarks. 
Consequently, we construct benchmarks for long dance generation using the AIST++ dataset. We first slice the music into 5-second clips, ensuring the latter 2.5-second clips of the previous music slice are the same as the former 2.5-second clips of the next music slice. Next, PAMD generates dance slices based on each of these 5-second clips. Then, dance slices with the same music part are combined via the weighted summation to generate the long dance sequence. According to Algorithm~\ref{alg:long}, we generate dance sequences of 7.5 seconds and 10 seconds durations, respectively. 

\subsection{Results of Dance Generation}

\noindent\textbf{Short-Term Generation: } 
The short-term generation is the standard setting that generates dances lasting 5 seconds (s). 
We compare PAMD with other existing dance generation methods, as shown in Table~\ref{tab:results}. 
Bailando~\cite{siyao2022bailando} and TM2D~\cite{gong2023tm2d} are dance generation methods based on VQ-VAE, while EDGE~\cite{tseng2023edge}, Lodge~\cite{li2024lodge} and the proposed PAMD are diffusion-based method. 

As shown in Table~\ref{tab:results}, our method PAMD outperforms all previous work in Beat Alignment Score, which is one of the most important metrics in choreography. This indicates that our method can generate dances that are better aligned with music.
Additionally, our method achieves a minimum on PFC, demonstrating that PAMD can generate physically more plausible motions.
Furthermore, our approach also outperforms previous methods in $FID_g$. This indicates that the dance quality generated by PAMD has achieved significant improvement in the geometric feature space. Next, compared to EDGE and Lodge, both using Diffusion model, PAMD performs best on $FID_k$ and $Div_k$, overcoming the poor quality and diversity of dances generated using Diffusion model in kinematic feature space.

\begin{figure}[tb]
	\centering
	\includegraphics[width=1\linewidth]{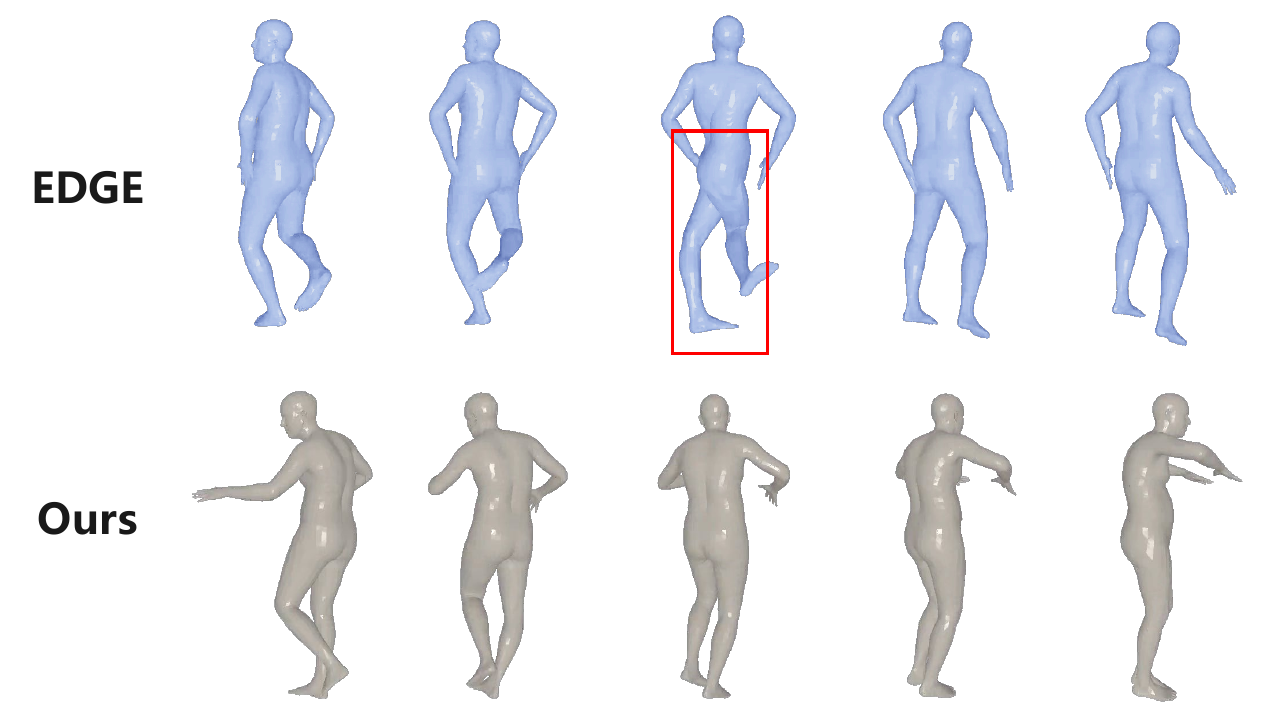}
	\vspace{-10pt}
	\caption{\textbf{Visualization of the Spinning Movement:} Grey and blue motions are dance motions generated by our method (PAMD) and EDGE, respectively. The motions generated by PAMD are smoother compared to those generated by EDGE, which displays a noticeable implausibility in the third frame.}
	\label{fig:PFC}
\end{figure}

\noindent\textbf{Long-Term Generation: }Long-term dance generation results are presented in Table~\ref{tab:long_results}. It can be seen that our approach still maintains a clear advantage in Beat Alignment Score metric in long dance generation. Even as the dance duration increases, the advantage of our method in Beat Alignment Score becomes more obvious. Specifically, for EDGE (w=2), the Beat Alignment Score is 0.26 at 7.5s and decreases to 0.25 at 10s. Conversely, for PAMD (w=2) (ours), the Beat Alignment Score is 0.30 at 7.5s and increases to 0.31 at 10s. 
Furthermore, as time increases, PAMD consistently outperforms EDGE in PFC metric. Additionally, we visualize the spinning movements in the dance, as shown in Figure~\ref{fig:PFC}. During the spinning process, EDGE exhibits implausible movements, while PAMD achieves a much smoother spinning motion. Furthermore, in the geometric feature space, $FID_g$ and $Div_g$ outperform EDGE, indicating that the long dances generated by PAMD better conform to predefined movement templates.

\subsection{Ablation Studies, Analysis, and Visualizations}
We perform ablation experiments to evaluate the impact of different parts in PAMD. The results are shown in Table~\ref{tab:ablations}.

\noindent\textbf{Effect of the MRFC: } 
Based on the results from  \textcircled{\scriptsize{1}} and Method \textcircled{\scriptsize{3}} presented in Table 3, it is observed that the inclusion of MRFC results in overall improvements in various performance metrics. Notably, the PFC score drops from 1.66 to 1.56, representing a reduction of approximately 6\%. This indicates that MRFC can further effectively assist the model in generating plausible motions.

We also compare our method with Lodge~\cite{li2024lodge}, which also incorporates a refine dance module. Table~\ref{tab:parameters} reveals that our approach not only has fewer parameters in terms of total model and refine module but also outperforms Lodge in Beat Alignment Score. 

\begin{table}[tbp]
	\caption{Comparison between Lodge and our approach.}
	\vspace{-5pt}
	\centering
	\begin{tabular}{l|c|cc}
		\hline
		\multicolumn{1}{l|}{\multirow{2}[0]{*}{Method}} & \multicolumn{1}{c|}{\multirow{2}[0]{*}{$BAS\uparrow $}} & \multicolumn{2}{c}{Parameters} \\
		\cline{3-4}
		&       & Method  & Refine Module \\
		\hline
		Lodge~\cite{li2024lodge} & 0.24 &  804.72 M& 4.51 M \\
		PAMD (ours) & \textbf{0.31}  &  207.44 M  & 1.50 M \\
		\hline
	\end{tabular}
	\label{tab:parameters}
\end{table}

\noindent\textbf{Effect of the PMC: } 
As shown in Table~\ref{tab:ablations}, it is evident that in the absence of PMC, Method \textcircled{\scriptsize{0}} performs the worst in both Beat Alignment score and PFC. However, when PMC is incorporated, Method \textcircled{\scriptsize{1}} outperforms Method \textcircled{\scriptsize{0}} in both metrics. Notably, in Method \textcircled{\scriptsize{4}}, where the model benefits from the combined influence of PMC, PMG, and MRFC, all metrics reach their optimal values. Specifically, compared to Method \textcircled{\scriptsize{0}}, the Beat Alignment Score increases from 0.27 to 0.31, a 15\% improvement, while the PFC decreases from 2.04 to 1.44, a 29\% reduction. This indicates that PMC not only effectively helps to generate dances that are better synchronized and more plausible but, when combined with PMG and MRFC, enables PAMD to perform even better, demonstrating the effectiveness of our strategy.

In addition, we also conduct an ablation experiment on the PMC module on physical metrics. According to Table~\ref{tab:PMC}, it can be seen that with the addition of PMC, the generated dances will be less likely to have the implausible movements of skating, floating and penetration.
\begin{table}[tbp]
	\centering
	\caption{The effectiveness of PMC.}
	\vspace{-5pt}
	\begin{tabular}{l|ccc}
		\hline
		Method &  Skating$\downarrow$  & floating$\downarrow$  & Penetration$\downarrow$  \\
		\hline
		Ours (w/ PMC)  & \textbf{0.179} & \textbf{0.567} & \textbf{0.369} \\
		Ours w/o PMC  & 0.221  & 0.641 & 0.374 \\
		\hline
	\end{tabular}
	\label{tab:PMC}
\end{table}

\noindent\textbf{Effect of the PMG: } 
We conduct experiments regarding the selection of Prior Motion and the results are shown in Table~\ref{tab:prior}. We consider two types of poses: a standard standing pose and a legs and feet open pose, both commonly observed across various dances. From the results, it is evident that the selection of the standard standing pose performs better in terms of Beat Alignment Score and exhibits superior quality and diversity in kinematic features.
\begin{table}[tbp]
	\centering
	\caption{The effectiveness of the prior motion.}
	\vspace{-5pt}
	\begin{tabular}{l|ccc}
		\hline
		Prior Motion &  $BAS\uparrow $ & $FID_k\downarrow$  & $Div_k\rightarrow$ \\
		\hline
		Standard standing  & \textbf{0.29}  & \textbf{39.47} & \textbf{5.21} \\
		Legs and feet open  & 0.28  & 45.60 & 4.88 \\
		\hline
	\end{tabular}
	\label{tab:prior}
\end{table}

Based on Method \textcircled{\scriptsize{1}} and Method \textcircled{\scriptsize{2}} presented in Table~\ref{tab:ablations}, it can be observed that the incorporation of PMG leads to a modest improvement across all performance metrics. Specifically, the Beat Alignment Score increases from 0.28 to 0.29, representing a 4\% improvement. And the PFC metric decreases from 1.66 to 1.62, reflecting a 2\% reduction. Additionally, both the FID and Diversity scores perform better. This suggests that the inclusion of PMG has a positive effect on our model performance.

\noindent\textbf{User Studies:}
To gain deeper insights into the real visual quality of our approach, we invite 11 participants to rate the generated dance performances. We present 21 pairs of comparison videos with the ground truth data from the AIST++ test set, which includes 10 pairs with a duration of 5 seconds, 8 pairs with a duration of 7.5 seconds, and 3 pairs with a duration of 10 seconds. We ask the participants:  \textit{“On a scale from 0 to 5, how would you rate this dance performance?”} The data is shown in Table~\ref{tab:user}. The majority (81.43\%) prefer the dances generated by our method. Notably, our method surpasses the ground truth with a 61.42\% winning rate.
\begin{table}[htbp]
	\centering
	\caption{The results of user study of dance generation.}
	\begin{tabular}{l|c}
		\hline
		Method & \multicolumn{1}{l}{Ours Wins} \\
		\hline
		Ground Truth  &  61.42\%\\
		\hline
		EDGE  &  81.43\%  \\
		\hline
	\end{tabular}
	\label{tab:user}
\end{table}

\begin{figure*}
	\centering
	\includegraphics[width=1\linewidth]{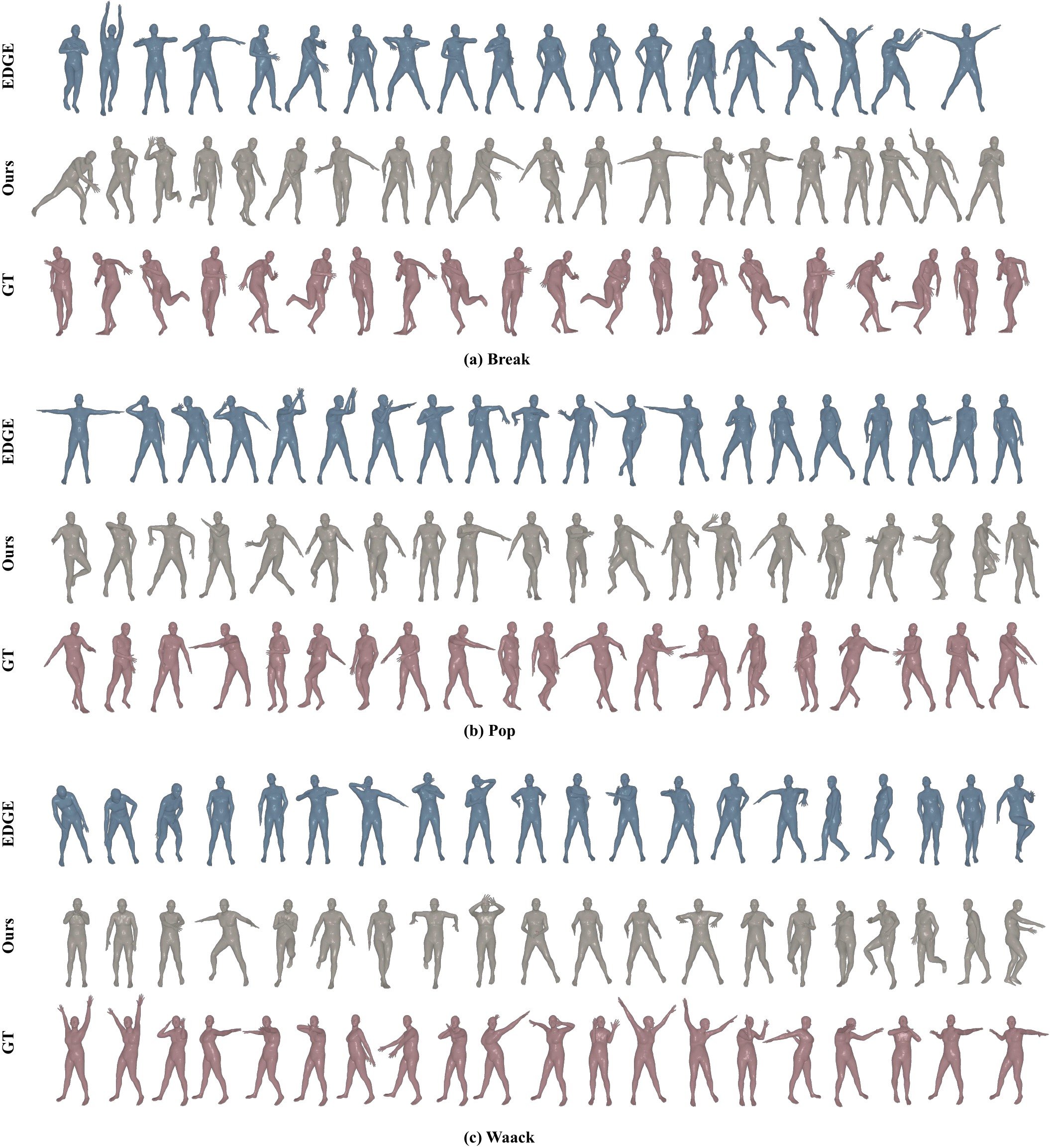}
	\caption{\textbf{Visualization of Long Dance Generation:} The dances with a duration of 10 seconds are shown in (a) break, (b) pop, and (c) waack. Each frame is sampled at 0.5-second intervals, resulting in 20 frames for each 10-second dance sequence. }
	\label{fig:vis}
\end{figure*}

\noindent\textbf{Visualization:}
We visualize the generated movements in Figure~\ref{fig:vis}. We generate 10-second dance sequences for Break, Pop, and Waack. The visualized results show that our method is capable of generating plausible long dances that are well aligned with the ground truth dances. More visualizations are available at: \url{https://mucunzhuzhu.github.io/PAMD-page/}.

\section{Conclusion}
In this paper, we study the problem of motion plausibility and propose a Plausibility-Aware Motion Diffusion (PAMD) for long dance generation. To enhance the physical plausibility of generating dances, PAMD introduces three modules in the diffusion model: Plausible Motion Constraint, Prior Motion Guidance, and Motion Refinement with Foot-Ground Contact. Through extensive experiments, the PAMD can generate long dances that not only align better with the conditioned music, but also exhibit higher physical plausibility. Ablation studies demonstrate the effectiveness and complementarity of these modules. Visualizations and user studies also demonstrate the visual plausibility of the generated long dance. This work has potential for music-driven human motion generation, automatic dance creation, and dance editing.

\ifCLASSOPTIONcaptionsoff
  \newpage
\fi
\bibliographystyle{IEEEtran}
\bibliography{IEEEbib}

\end{document}